\documentclass[letterpaper, 10 pt, conference]{ieeeconf}  % Comment this line out if you need a4paper

\IEEEoverridecommandlockouts                    % This command is only needed if 
\usepackage{graphics} % for pdf, bitmapped graphics files
\usepackage{amsmath} % assumes amsmath package installed
\usepackage{amssymb}  % assumes amsmath package installed
\usepackage[linesnumbered,ruled,vlined]{algorithm2e}
\usepackage{caption}
\usepackage{subcaption}
\usepackage[table]{xcolor}
\usepackage{tabu}
\usepackage{booktabs}       % professional-quality tables
\usepackage{amsfonts}       % blackboard math symbols
\usepackage{nicefrac}       % compact symbols for 1/2, etc.
\usepackage{multirow}
\usepackage{adjustbox}

\title{\LARGE \bf
Hierarchical Planning Through\\
Goal-Conditioned Offline Reinforcement Learning 
}

\author{Jinning Li, Chen Tang, Masayoshi Tomizuka, Wei Zhan% <-this % stops a space
\thanks{J. Li, C. Tang, M. Tomizuka, and W. Zhan are with the Department of Mechanical Engineering, University of California, Berkeley, CA 94720, USA
        {\tt\small 	\{jinning\_li, chen\_tang, tomizuka, wzhan\}@berkeley.edu}. Corresponding author: Jinning Li.}%
\thanks{This work is supported by Denso International America, Inc.}
}

\begin{document}

\maketitle
\thispagestyle{empty}
\pagestyle{empty}

%%%%%%%%%%%%%%%%%%%%%%%%%%%%%%%%%%%%%%%%%%%%%%%%%%%%%%%%%%%%%%%%%%%%%%%%%%%%%%%%
\begin{abstract}

Offline Reinforcement learning (RL) has shown potent in many safe-critical tasks in robotics where exploration is risky and expensive. However, it still struggles to acquire skills in temporally extended tasks.  In this paper, we study the problem of offline RL for temporally extended tasks. We propose a hierarchical planning framework, consisting of a low-level goal-conditioned RL policy and a high-level goal planner.  The low-level policy is trained via offline RL. We improve the offline training to deal with out-of-distribution goals by a perturbed goal sampling process.
The high-level planner selects intermediate sub-goals by taking advantages of model-based planning methods. It plans over future sub-goal sequences based on the learned value function of the low-level policy. We adopt a Conditional Variational Autoencoder to sample meaningful high-dimensional sub-goal candidates and to solve the high-level long-term strategy optimization problem. We evaluate our proposed method in long-horizon driving and robot navigation tasks.
Experiments show that our method outperforms baselines with different hierarchical designs and other regular planners without hierarchy in these complex tasks. 

\end{abstract}


%%%%%%%%%%%%%%%%%%%%%%%%%%%%%%%%%%%%%%%%%%%%%%%%%%%%%%%%%%%%%%%%%%%%%%%%%%%%%%%%
\section{Introduction} \label{sec:intro}

Reinforcement learning (RL) has been widely applied for a broad range of tasks in robotics. However, it remains challenging to solve those complex tasks with extended temporal duration with RL, especially for safety-critical applications where exploration is risky and expensive (e.g., autonomous driving). To avoid risky exploration, offline RL has drawn growing research attention recently \cite{fu2020d4rl, offline-rl-tutorial, cql, li2022dealing}, for its ability to train RL policies from static offline datasets. Many prior works have investigated offline RL algorithms in the area of robotics and autonomous driving~\cite{pmlr-v164-sinha22a, kumar2021workflow, singh2020cog}. However, no prior works have deliberately investigated offline RL algorithms for temporally extended tasks. In this work, we aim to study this problem, which is an important step towards the applications of RL to navigate intelligent agents in complex and safety-critical environment. 

To deal with extended temporal duration, a promising solution is to adopt a hierarchical framework, with a low-level policy controlling the agent to accomplish short-term tasks, while receiving supervision from a high-level module reasoning about long-term strategy. The high-level module could be a model-free RL policy \cite{dietterich1998maxq, nachum2018data, li2021safe} or a planning module \cite{nasiriany2019planning}. We are particularly interested in \cite{nasiriany2019planning}, which combines a low-level goal-conditioned RL policy \cite{Kaelbling93learningto, pmlr-v37-schaul15} and a high-level model-based planning module guiding the goal-reaching policy with a sub-goal sequence. The high-level planning takes the advantages of model-based approaches \cite{punjani2015deep, berkenkamp2017safe} to handle long-horizon tasks by composing behaviors. We found it particularly suitable for offline setting because it only requires learning a short-horizon goal-reaching policy from offline dataset, which is much easier and more data-efficient than directly learning an end-to-end policy solving the entire task. 

In this work, we propose a hierarchical planning framework through goal-conditioned offline reinforcement learning. We leverage Hindsight Experience Replay (HER) \cite{her} to synthesize goal-conditioned episodes from static datasets. Afterwards, we train the policy with state-of-the-art model-free offline RL algorithms. To deal with the distributional shift caused by out-of-distribution (OOD) goals, we propose to generate noisy unreachable goals by a perturbed goal sampling process. By incorporating those noisy goals into training, it informs the high-level planner to avoid OOD goals during online execution. The high-level sub-goal planner plans over intermediate sub-goals for the low-level policy. Specifically, it solves an optimization problem based on the goal-conditioned value function of the low-level policy online. To ensure efficient online computation, we train a Conditional Variational Autoencoder (CVAE)~\cite{cvae} to propose goal sequence candidates that are feasible and reasonable.

We evaluate our proposed framework for planning in various domains, specifically, robot navigation tasks and driving scenarios. Experiment results show that our framework is superior to the policy trained by regular offline reinforcement learning methods without a hierarchical design. The trained value function can quantify the quality of different goal candidates in the high-level goal planner. Also, our framework can be generalized to complex driving scenes with lower collision rates than baselines. 

%%%%%%%%%%%%%%%%%%%%%%%%%%%%%%%%%%%%%%%%%%%%%%%%%%%%%%%%%%%%%%%%%%%%%%%%%%%%%%%%
\section{Related Works}

Hierarchical framework has shown promising in dealing with long-horizon tasks.
Conventional hierarchical RL decomposed the action space directly~\cite{duan2020hierarchical, chen2019attention}.
Recent works have proposed different architecture designs with novel high-level and low-level components, e.g., high-level graph search-based planner~\cite{eysenbach2019search}, two-phase training~\cite{mandlekar2020learning}, mixed imitation learning and reinforcement learning policies~\cite{gupta2019relay, mandlekar2020iris}, and high-level latent primitive extraction~\cite{ajay2020opal}. They typically require training multiple policies online, or the task to be repetitive. In contrast, our proposed framework only requires training one RL low-level policy from static offline data. Also, the value function used by the high-level planner corresponds to the short Markov decision process (MDP) between consecutive sub-goals instead of the whole MDP as in previous works~\cite{mandlekar2020iris}. Consequently, our method only requires estimation of the expected future return over the short horizon between sub-goals, instead of the entire episode length. Plus, our method can consider multiple look-ahead timesteps instead of one as in previous works~\cite{mandlekar2020iris,ajay2020opal}, increasing the ability to reason about long-term strategies.

Our high-level planner is similar to the one in~\cite{nasiriany2019planning}, which is developed for a finite-horizon goal-conditioned MDP with a single objective of goal reaching.
In contrast, ours generalizes it to MDPs that are not goal-conditioned, so that we can consider criteria besides reachability (e.g., comfort, safety). In \cite{nasiriany2019planning}, a variational autoencoder (VAE) is adopted to model high-dimensional image observation for efficient sub-goal sampling during online optimization. The sub-goals at different timesteps are modeled independently, inducing unnecessary noise into the high-level optimization. We instead use a CVAE to sequentially sample sub-goal sequences which improves the optimization performance. Most importantly, our framework is developed under the offline setting, whereas the planner in~\cite{nasiriany2019planning} requires learning the low-level policy and value function with online exploration.

\section{Problem Formulation}

We consider a MDP represented by a tuple $\mathcal{M} = (\mathcal{S}, \mathcal{A}, \mathcal{P}, r_\text{env}, \gamma)$, where $\mathcal{S}$ is the state space, $\mathcal{A}$ is the action space, $\mathcal{P}$ is the transition function, $r_\text{env}$ is the reward function, and $\gamma$ is the discount factor. An important component when using RL to solve a MDP is the value function under a policy. Value function estimation tends to be accurate when predicting at points in the neighborhood of the present observation~\cite{when-to-trust-model}. It is then reasonable to break a big task into small pieces, so that the policy only needs the corresponding value function to predict values at close and accurate goal points. Based on this insight, we solve the MDP with a hierarchical policy where a high-level planner is responsible for sub-goal selection, and a low-level goal-conditioned policy is responsible for generating executable actions.

The goal planner operates at a coarse time scale, generating a sub-goal at every $N$ time steps. We define the time steps at which the goal planner outputs a sub-goal as \emph{high-level time steps}, denoted by $t_i$, with $t_{i+1} - t_{i} = N$. Given a planned sub-goal at $t_{i}$, the low-level policy generates actions at each time step between $t_i$ and $t_{i+1}$, controlling the agent to reach the desired sub-goal at $t_{i+1}$. Formally, the low-level goal-conditioned policy is denoted by $\pi(\mathbf{a}_{t_{i, j}} | \mathbf{s}_{t_{i, j}}, \mathbf{g}_{t_{i+1}})$ for $j=0, 1, ..., N-1$, where $t_{i} = t_{i, 0}$, and $\mathbf{g}_{t_{i+1}}$ is the sub-goal for $t_{i+1}$ generated at the previous high-level time step. The low-level policy essentially solves a short-horizon goal-reaching MDP defined between two consecutive high-level time steps. In the next section, we will show that the planning problem can be formulated as an optimization problem with an objective function defined with the value function of the goal-reaching MDP.

%%%%%%%%%%%%%%%%%%%%%%%%%%%%%%%%%%%%%%%%%%%%%%%%%%%%%%%%%%%%%%%%%%%%%%%%%%%%%%%%
\section{Hierarchical Goal-Conditioned Offline Reinforcement Learning}

\begin{figure}[t]
    \centering
    \includegraphics[width=0.35 \textwidth]{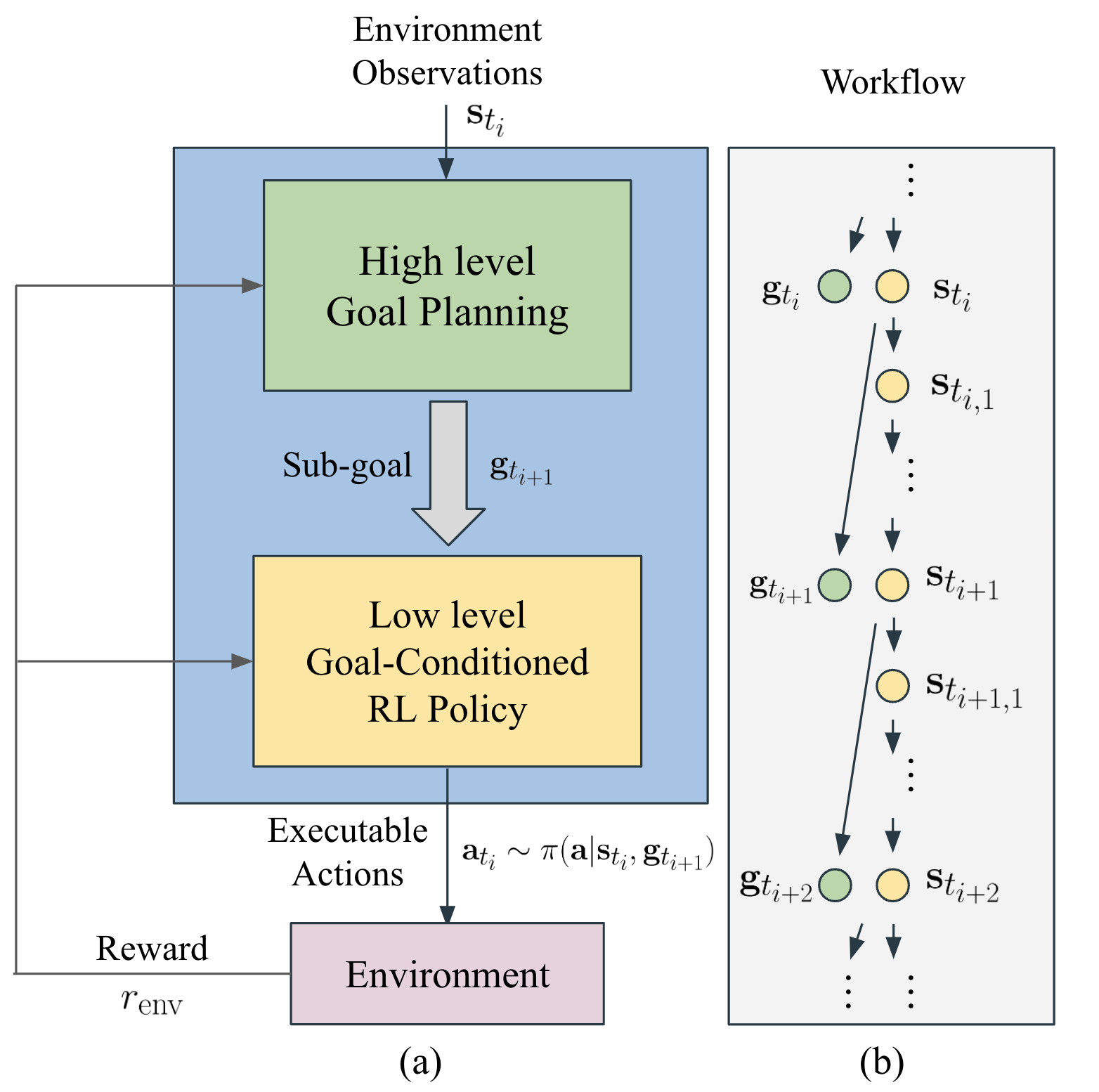}
    \caption{(a) The hierarchical goal-conditioned offline reinforcement learning framework.
    (b) The workflow of the proposed framework.}
    \label{fig:flow}
    \vspace{-0.2cm}
\end{figure}

We now present the proposed hierarchical planning framework shown in Fig.~\ref{fig:flow} by answering the following questions: 1) How the planner generates sub-goals; 2) How to combine the sub-goal planner and the goal-conditioned policies; and 3) How to train goal-conditioned policies by offline reinforcement learning algorithms. 

\subsection{Generating High-Level Sub-Goals}

While the low-level policy only requires a single sub-goal, we let the planner optimize over multiple sub-goals into the future to ensure an optimal long-term strategy. It can be formulated as a constrained optimization problem: maximizing the cumulative reward over sub-goals for multiple future steps with a constraint to force the sub-goals to be reachable by the low-level policy. The solution is a sequence of $H$ sub-goals for $H$ future high-level time steps. We select the first one to execute and repeat the planning process at the next time step.

Formally, The constrained optimization problem is formulated as follows:
\begin{gather}
\label{eq:constraint-opt}
\begin{aligned}
    & \max_\mathbf{g} V^\pi_\text{env}(\mathbf{s}_{t_0}, t_0) 
    + \sum_{i=1}^{H} V^\pi_\text{env}(\mathbf{g}_{t_i}, t_i)\\
    & \text{s.t.} \;\; V^\pi_\text{TDM}(\mathbf{s}_{t_0}, \mathbf{g}_{t_1}, t_0) 
    + \sum_{i=1}^{H} V^\pi_\text{TDM}(\mathbf{g}_{t_i}, \mathbf{g}_{t_{i+1}}, t_{i}) \geq 0,
\end{aligned}
\end{gather}
where  $\mathbf{g} = \begin{bmatrix} \mathbf{g}_{t_1}^\top, \dots, \mathbf{g}_{t_H}^\top \end{bmatrix}^\top$ is the sub-goal sequence of interest, $\mathbf{s}_{t_0}$ is the initial state, and
\begin{gather*}
    \begin{aligned}
    & V^\pi_\text{env}(\mathbf{s}_{t_i}, t_i) = \mathbb{E}\left[ \sum_{j=1}^{N} r_\text{env}(\mathbf{s}_{t_{i, j}}, \mathbf{a}_{t_{i, j-1}}) | \pi \right], \\
    & V^\pi_\text{TDM}(\mathbf{s}_{t_i}, \mathbf{g}_{t_{i+1}}, t_i) = \mathbb{E}\left[ \sum_{j=1}^{N} r_\text{TDM}(\mathbf{s}_{t_{i,j}}, \mathbf{g}_{t_{i+1}}, t_{i,j}) | \pi  \right],
    \end{aligned}
\end{gather*}
where $r_\text{env}$ is the reward function defined in the original MDP environment, and $r_\text{TDM}$ is the auxiliary goal-conditioned reward function. $r_\text{TDM}$ is defined through a temporal difference model (TDM) that is commonly adopted in prior goal-conditioned policy learning works~\cite{nasiriany2019planning, her, goal-IL}. 
The TDM reward function is in the form:
\begin{gather*}
    r_\text{TDM} (\mathbf{s}_\tau, \mathbf{g}_t, \tau) = -\delta (\tau=t) d(\mathbf{s}_\tau, \mathbf{g}_t),
\end{gather*}
where $\delta(\cdot)$ is the indicator function, and $d$ is a distance function specified in tasks. By definition, the TDM reward is always non-positive. The constraint in Problem~(\ref{eq:constraint-opt}) forces the expected cumulative reward $r_\text{TDM}$ to be non-negative. Consequently, a feasible $\mathbf{g}$ has $r_\text{TDM}=0$ for all the time steps, which means that $\mathbf{g}$ is guaranteed to be reachable.

The optimal solution of Problem~(\ref{eq:constraint-opt}) can be obtained by solving the following max-min optimization problem, where the objective is the Lagrangian function of Problem~(\ref{eq:constraint-opt}):
\begin{gather}
    \max_\mathbf{g} \min_{\beta\geqslant 0} V^\pi(\mathbf{s}_{t_0}, \mathbf{g}_{t_1}, t_0)
    + \sum_{i=1}^{H} V^\pi(\mathbf{g}_{t_i}, \mathbf{g}_{t_{i+1}}, t_{i}),
    \label{eq:minmax}
\end{gather}
where 
\begin{align}
    V^\pi(\mathbf{s}_{t_i}, \mathbf{g}_{t_{i+1}}, t_{i})
    &= V^\pi_\text{env}(\mathbf{s}_{t_i}, t_i) + \beta V^\pi_\text{TDM}(\mathbf{s}_{t_i}, \mathbf{g}_{t_{i+1}}, t_i) \nonumber\\
    &= \mathbb{E}\left[ \sum_{j=1}^{N} r_\text{env} + \beta r_\text{TDM} | \pi \right]. \label{eqn:value-function}
\end{align}

The augmented value function $V^\pi$ is essentially the value function of the low-level policy $\pi$ regarding an augmented reward function $r_g$, which is the weighted sum of the original environmental reward and the auxiliary TDM reward:
\begin{gather*}
    r_g(\mathbf{s}_\tau, \mathbf{a}_\tau, \mathbf{g}_t, \tau) = 
    r_\text{env}(\mathbf{s}_\tau, \mathbf{a}_\tau) + \beta r_\text{TDM}(\mathbf{s}_\tau, \mathbf{g}_t, \tau).
\end{gather*}

We may apply dual gradient descent to iteratively update $g$ and the Lagrangian multiplier $\beta$. However, our hierarchical planning framework requires solving the optimization problem timely online. To this end, we consider the Lagrangian relaxation of Problem~(\ref{eq:minmax}), which is obtained by specifying a prefixed value of $\beta$ to relax the problem into an unconstrained maximization problem, where the multiplier $\beta$ is essentially a trade-off factor between the objective and the regularization term:
\begin{gather}
    \max_\mathbf{g} V^\pi(\mathbf{s}_{t_0}, \mathbf{g}_{t_1}, t_0)
    + \sum_{i=1}^{H} V^\pi(\mathbf{g}_{t_i}, \mathbf{g}_{t_{i+1}}, t_{i}).
    \label{eq:uncon-opt}
\end{gather}

We follow the convention to use the cross-entropy method (CEM) as the optimizer~\cite{cem, nasiriany2019planning} to solve Problem~(\ref{eq:uncon-opt}). In this paper, we focus on tasks with high-dimensional images as observations. To this end, we need to sample meaningful images as valid goal states from the state space $\mathcal{S}$. In general, we do not have an explicit high-dimensional bound between meaningful images and invalid white noise, but it is possible to find a method that implicitly encourages valid samples. A natural method is modeling the goal distribution with a generative model such as VAE~\cite{nasiriany2019planning}. In particular, we adopt a CVAE~\cite{cvae} in this work. 

The CVAE model consists of an inference mapping, i.e., an encoder $E_\mu (\mathbf{z} | \mathbf{g}_{t_i}, \mathbf{g}_{t_{i+1}})$, and a generative mapping, i.e., a decoder $D_\nu (\mathbf{g}_{t_{i+1}} | \mathbf{g}_{t_i}, \mathbf{z})$. The inference network maps a present goal state $\mathbf{g}_{t_i} \in \mathcal{S}$ to a latent state $\mathbf{z} \in \mathcal{Z}$ condition on its next goal $\mathbf{g}_{t_{i+1}}$, where $\mathcal{Z}$ is the latent space. The generative mapping, which conditions on the original input $\mathbf{g}_{t_i}$, maps the latent state $\mathbf{z} \in \mathcal{Z}$ to the next goal $\mathbf{g}_{t_{i+1}}$. The latent space $\mathcal{Z}$ is often low-dimensional, and the latent state $\mathbf{z}$ follows a specified prior distribution $p(\mathbf{z})$. By decoding goal states from latent variables sampled from the prior $p(\mathbf{z})$, we essentially sample goal states from a distribution approximating the goal distribution in the dataset. The sampled images are then more likely to be meaningful and correspond to in-distribution goal states. Different from the VAE used in~\cite{nasiriany2019planning}, conditioning on the present state image reduces unnecessary variance. The sampled goals are those that are more likely given the current state. It improves sampling efficiency and leads to better optimization performance. 

The objective function can then be optimized over the latent representation $\mathbf{z} = \begin{bmatrix} \mathbf{z}_{t_1}^\top, \dots, \mathbf{z}_{t_H}^\top \end{bmatrix}^\top$ of the goal images $\mathbf{g} = \begin{bmatrix} \mathbf{g}_{t_1}^\top, \dots, \mathbf{g}_{t_H}^\top \end{bmatrix}^\top$, where $\mathbf{g}_{t_{i+1}} = D_\nu (\mathbf{g}_{t_{i}}, \mathbf{z}_{t_{i+1}}) $ is the reconstructed goal from sampled latent state $\mathbf{z}_{t_{i+1}}$ conditioned on the previous goal $\mathbf{g}_{t_{i}}$. An additional regularization term $-\lambda \log p(\mathbf{z})$ is added to the objective function to penalize $\mathbf{z}$ with low prior probability.

\subsection{Goal-Conditioned Offline Reinforcement Learning}
To train the low-level policy from static dataset, we aim to combine goal-conditioned training techniques with offline reinforcement learning. In particular, we need to train not only the policy $\pi$, but also the value function $V^\pi$ in Eqn.~(\ref{eqn:value-function}). 

\subsubsection{Relabeling the Expert}

The pre-collected datasets to train the goal-conditioned policies only include state and action transition pairs and corresponding rewards. 
There are no goals defined in training datasets originally, so we must create a goal for each sampled state-action pair during training. 
The trajectories are valid ones to reach any states within themselves, which makes it reasonable to use any state within each trajectory as its goal.

Therefore, following the hindsight experience replay~\cite{her}, if we get a sampled pair $(\mathbf{s}_{t_{i,j}}^k, \mathbf{a}_{t_{i,j}}^k, \mathbf{s}_{t_{i,j+1}}^k, r_{\text{env}, t_{i,j}}^k)$ from the $k$-th trajectory in the training dataset, then we will relabel its goal as $\mathbf{g}_{t_{i+1}}^k = \mathbf{s}_{\tau}^k$ where $\mathbf{s}_{\tau}^k$ is a random future state within the whole trajectory. The new reward will be relabeled as:
\begin{equation}
    r_{g, t_{i,j}}^k \leftarrow r_{\text{env}, t_{i,j}}^{k} + \beta r_{\text{TDM}, t_{i,j}}^k. \label{eqn:reward-update}
\end{equation}
The TDM reward is computed regarding the labeled goal $\mathbf{g}_{t_{i+1}}^k$. The new pair $(\mathbf{s}_{t_{i,j}}^k, \mathbf{a}_{t_{i,j}}^k, \mathbf{s}_{t_{i,j+1}}^k, \mathbf{g}_{t_{i+1}}^k, r_{g, t_{i,j}}^k)$ will be used as the training data to update the goal-conditioned value function and policy. 

\begin{algorithm}[t]
\SetAlgoLined
    \textbf{Initialize}: A Q-network $Q_\theta$ parametrized by $\theta$, A target network $Q_{\Bar{\theta}} = Q_\theta$ parametrized by $\Bar{\theta}$, a policy network $\pi_\varphi$ parametrized by $\varphi$, an encoder $E_{\mu}$ and a decoder $D_{\nu}$ for the CVAE, a training dataset $\mathcal{D}$\;
    \For{step $c$ in range($0$, $C$)}{
        Sample a batch of $b$ states $\mathbf{s}$ from the dataset $\mathcal{D}$\;
        Update $\mu$ and $\nu$ according to the CVAE objective\;
    }
    \For{step $m$ in range($0$, $M$)}{
        % Sub-goal planner Training
        Sample a pair $(\mathbf{s}_{t_{i,j}}^k, \mathbf{a}_{t_{i,j}}^k, \mathbf{s}_{t_{i,j+1}}^k, r_{\text{env}, t_{i,j}}^k)$ \;
        Sample a future state $\mathbf{s}_\tau^k$ within the $k$-th trajectory and add noise with a probability $\eta$ to obtain the goal $\mathbf{g}_{t_{i+1}} = D_\nu (E_\mu (\mathbf{s}_\tau^k)+\varepsilon, \mathbf{s}_{\tau-N}^k)$\;
        Relabel the reward following Eqn.~(\ref{eqn:reward-update})\;
        Update $Q_\theta$ with the CQL policy evaluation step and learning rate $\epsilon_\theta$:
        $\theta_m \leftarrow \theta_{m-1} + \epsilon_\theta \nabla_\theta J(\theta)$\;
        Update $\pi_\varphi$ according to the soft actor-critic style objective and learning rate $\epsilon_\varphi$:
        \begin{align*}
            \varphi_m \leftarrow \varphi_{m-1} + & \epsilon_\varphi \mathbb{E}_{\mathbf{s}\sim d^{\pi_\beta}(\mathbf{s}), \mathbf{a}\sim \pi_\varphi(\mathbf{a})}[Q_\theta(\mathbf{s}, \mathbf{a}, \mathbf{g}) \\
            - & \log \pi_\varphi(\mathbf{a}|\mathbf{s}, \mathbf{g})];
        \end{align*}\\
        \If{$ m $ \text{mod target\_update} == $0$}{
            Soft Update the target network $\Bar{\theta}_{m} \leftarrow (1-\tau)\Bar{\theta}_{m-1} + \tau \theta_{m-1}$
        }
    }
    \caption{Training Procedure of Hierarchical Goal-Conditioned Offline Reinforcement Learning}
    \label{alg:train}
\end{algorithm}

\subsubsection{Robust to Distributional Shift} \label{sec:robust-to-ood}
Many prior works on offline RL have dealt with the infamous distributional shift problems~\cite{cql, li2022dealing}. Under the goal-conditioned setting, we need to additionally handle OOD goals. It is similar to the problem of OOD states. However, when dealing with OOD states, existing works require a dynamic model to estimate state uncertainty \cite{li2022dealing}, which is extremely difficult to learn in high-dimensional observation space. Alternatively, we propose a new method to solve the distributional shift problem caused by OOD goals.
Concretely, we perturb the goal in the sampled data point by a noise with a probability $\eta$, and then the reward $r_{g, t_{i,j}}^k$ is relabeled as the new noisy data point. In this way, we penalize the value at OOD goals, because a $\eta$-portion of the goals in the sampled data batch become noisy and were never achieved in the dataset. Thus, the corresponding TDM rewards will always be negative. Hence, the high-level goal planner using the same value function will be encouraged to avoid those low value areas at OOD goals, and thus it will implicitly affect the low-level policy during test time. 

\subsection{Practical Implementations}

We now integrate the high-level goal planner and the low-level goal-conditioned policy to form a practical implementation, which we call the Hierarchical Goal-Conditioned offline reinforcement learning (HiGoC) framework. Given a static offline dataset containing $K$ expert trajectories:
$$\mathcal{D}=\left\{ \{(\mathbf{s}_{t_{i,j}}^k, \mathbf{a}_{t_{i,j}}^k, \mathbf{s}_{t_{i,j+1}}^k, r_{\text{env}, t_{i,j}}^k)\}_{i,j}, k=1, \dots, K \right\},$$ 
the training procedure is shown in Alg.~\ref{alg:train}. A CVAE is first trained to model the goal distribution in the dataset. Then, the low-level goal-conditioned policy and the corresponding value function is trained through Conservative Q-Learning (CQL)~\cite{cql}, which is customized for the goal-conditioned settings. 
In the test phase, we follow Alg.~\ref{alg:test} to query the hierarchical planner to control the agent. 

\begin{algorithm}[t]
\SetAlgoLined
    \For{high-level step $t_i$ in range($0$, $T$)}{
        Solve the optimization problem in Eqn.~(\ref{eq:uncon-opt}) for a sub-goal $\mathbf{g}_{t_{i+1}}$\;
        \For{low-level step $t_{i,j}$, with $j$ in range($0$, $N$)}{
        Sample $\mathbf{a}_{t_{i,j}} \sim \pi_\varphi(\mathbf{a} \; | \; \mathbf{s}_{t_{i,j}}, \mathbf{g}_{t_{i+1}})$\;
        Execute the action $\mathbf{a}_{t_{i,j}}$\;
        }
    }
    \caption{Hierarchical Goal-Conditioned Planner}
    \label{alg:test}
\end{algorithm}

\section{Experiments}
In this section, we present the results of our experiments with HiGoC. We intend to mainly answer the following questions with our experiments:

\begin{itemize}
    \item Does HiGoC have better performance than the end-to-end RL policy and other hierarchical baselines? 
    %\item How does the quality of the training datasets affect the performance of the learned policy?
    \item Does longer look-ahead horizon $H$ in the goal-selection optimization problem lead to better performance?
    \item Does the value function effectively encodes the optimality of different goal points? 
    \item Do the perturbed goal sampling process and the CVAE improve the performance of the hierarchical planner?
    %\item Can we apply our method to solve driving tasks in more realistic and complex scenarios?
\end{itemize}

We investigate those questions in simulated autonomous driving scenarios and robot navigation tasks. All experiments are evaluated and averaged over five random seeds and we report error bars corresponding to one standard deviation. 

\subsection{Experiment Settings}

\subsubsection{The CARLA Simulator}

For autonomous driving scenarios, we evaluate our method in the CARLA simulator~\cite{fu2020d4rl, carla, chen2019model, codevilla2018end}, in particular, the driving tasks specified in \cite{chen2019model}. In these tasks, the ego agent drive in a virtual urban town where the reward is given by the simulator evaluating the safety and efficiency of driving. There are 30 obstacle vehicles in total running in the whole map, so that we can simulate a realistic environment with intense interactions with surrounding vehicles. In our experiments, we first confine the evaluation in a local map containing a roundabout in the virtual town (Sec.~\ref{sec:small-map}). Afterwards, we extend the experiment to the whole town to evaluate the ability of our algorithm in building reliable policies in more general and realistic scenes (Sec.~\ref{sec:big-map}). We choose the ``Town03'' map from the CARLA simulator for evaluation. 

In our experiments, static pre-collected training datasets are required to train the planner. To collect training data, an expert is trained using soft actor-critic (SAC)~\cite{sac} method with the default reward function in each environment.
We then execute the expert in the corresponding environment to record trajectories. After the collection of training datasets, we train HiGoC in the offline setting. During the test time, we drive the agent with the trained policies in the simulator.

We compose datasets with different levels of quality, namely ``medium'' and ``expert'' datasets. ``Medium'' datasets are collected by first training the expert using SAC, early stopping the training, and recording the trajectories with the half-trained expert in the environment. ``Expert'' datasets are collected with fully trained experts. 

\subsubsection{Antmaze in D4RL}

In addition to the driving task, we evaluate our method on robot navigation tasks from offline RL benchmarks, which are the antmaze tasks in the D4RL dataset~\cite{fu2020d4rl}. The dataset mainly consists of trajectories collected with undirected navigation of the simulated ant robot. The agent has to learn from the dataset to solve specific point-to-point navigation tasks. The task can be categorized into three difficulty levels: simple (i.e., umaze), medium, and difficult (i.e., large).

\subsubsection{Variants and Baselines} \label{sec:variant}
We mainly compare HiGoC with two baselines: a) CQL~\cite{cql}, which is one of the state-of-the-art offline RL algorithms for end-to-end policy training; b) IRIS~\cite{mandlekar2020iris}, which is also a hierarchical framework learned from offline dataset, with a high-level offline RL goal planner and a low-level imitation learning policy. For the antmaze experiments, we include another approach for comparison, OPAL~\cite{ajay2020opal}, which is another hierarchical framework with a high-level strategy planner and a low-level primitive controller. Also, we compare different settings of HiGoC: 
\begin{itemize}
    \item {\bf Different look-ahead horizon $H$}: Longer horizon enables the planner to better optimize long-term behavior. Meanwhile, errors of the value function and the CVAE model accumulate along the planning horizon. We are curious about how these two factors will trade off and affect the overall performance. We use the notation ``$H$-step'' to distinguish planners with different $H$. For instance, ``$2$-step'' refers to a planner with $H=2$.
    \item {\bf Different goal-sampling period $N$}: The goal-sampling period determines the episode length of the low-level goal-reaching MDP. As a result, it affects the accuracy of the learned value function. We specify the variants with different sampling period as ``HiGoC-$N\Delta t$''. For instance, ``HiGoC-$0.4$s'' refers to a variant with $N=4$ since the sampling time of the simulator is $0.1s$. 
    \item {\bf Variant without goal perturbation}: The last variant we study is the one without goal sampling perturbations during training, denoted by ``HiGoC-no noise''. It has a goal-sampling time of $0.4$s. By comparing it with the variant ``HiGoC-$0.4$s'', we would like to verify if the proposed method can effectively mitigate the issue of OOD goals and lead to better performance. 
\end{itemize}

\subsubsection{Evaluation Metric}

The main evaluation metric is the normalized score (NS)~\cite{fu2020d4rl}. The NS is defined as:
$$\texttt{NS = 100 * $\dfrac{\texttt{score - random score}}{\texttt{expert score - random score}}$}.$$
The score of each variant is the cumulative reward during test time after the whole training process finishes, namely, after $500$K gradient steps in CARLA and $1$M in antmaze. For the complex whole-town driving task, we also report the collision rate to give the audience a straightforward impression on the driving skill of the trained agents. We roll out 100 episodes, and compute the ratio of trajectories containing collisions.

%A score of $0$ corresponds to an agent that randomly chooses actions in the environment, whereas a score of $100$ corresponds to expert one that collects the dataset.

\subsection{Performance Comparison in CARLA} \label{sec:small-map}
We first report the results of a comprehensive study on the small local map, comparing all the different settings listed in Sec. \ref{sec:variant}. The results are summarized in Table \ref{tab:results}. 

\subsubsection{The Hierarchical Structure} \label{sec:hierarchical-struct}

\begin{figure*}[t]
    \centering
    \includegraphics[width=0.6\textwidth]{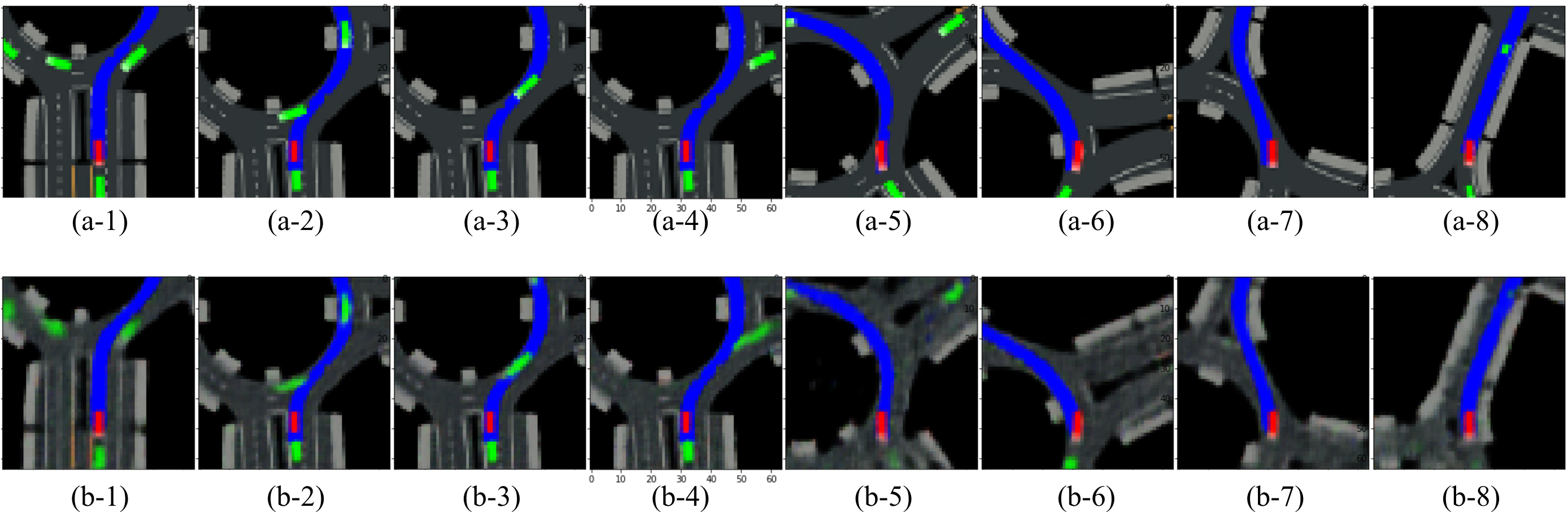}
    \caption{Visualization of the trajectory in the local map of a roundabout. We skip $10$ frames in between these sampled images for brevity.
    (a) The actual observation images; 
    (b) The corresponding goals selected by the goal planner with CVAE.}
    \label{fig:ra-vis}
\end{figure*}

\begin{table*}[h]
    \centering
  \begin{adjustbox}{max width=\textwidth}
    \begin{tabular}{l|c|c|c|c|c|c}
        \toprule
        \textbf{Dataset} & \textbf{Look-ahead} & \textbf{HiGoC-$0.4$s} & \textbf{HiGoC-$2.0$s} & \textbf{HiGoC-no noise} & \textbf{IRIS} & \textbf{CQL} \\
         \midrule
         \multirow{4}{4em}{\textbf{Medium}}
         & 1-step & $93.5\pm9.4$ & $98.4\pm10.3$ & $90.8\pm6.9$ & $83.3\pm5.3$ & $88.7\pm6.3$ \\
         & 3-step & $93.7\pm10.1$ & $95.5\pm11.4$ & $91.2\pm9.8$ & $87.2\pm4.5$ & - \\
         & 5-step & $96.2\pm13.5$ & $94.9\pm10.2$ & $95.5\pm14.7$ & $91.5\pm3.1$ & - \\
         & 7-step & $99.4\pm12.9$ & $92.7\pm15.8$ & $93.1\pm13.3$ & $90.1\pm7.8$ & - \\
         \midrule
         \midrule
         \multirow{4}{4em}{\textbf{Expert}} 
         & 1-step & $92.8\pm6.5$ & $97.4\pm7.2$ & $93.4\pm6.9$ & $84.1\pm4.7$ & $90.5\pm4.1$ \\
         & 3-step & $94.2\pm6.8$ & $96.5\pm7.7$ & $93.7\pm7.3$ & $92.3\pm6.0$ & - \\
         & 5-step & $97.9\pm8.1$ & $95.8\pm9.4$ & $94.6\pm8.1$ & $87.5\pm5.2$ & - \\
         & 7-step & $98.4\pm9.4$ & $91.6\pm11.1$ & $95.3\pm10.5$ & $89.9\pm4.8$ & - \\
         \bottomrule
    \end{tabular}
    \end{adjustbox}
    \caption{Normalized scores of all variants when trained with the ``Medium'' and ``Expert'' quality dataset in the local map.}
    \label{tab:results}
    \vspace{-0.3cm}
\end{table*}

We first compare HiGoC with the end-to-end policy baseline to evaluate the benefit of the hierarchical structure. From Table.~\ref{tab:results}, we can see that the performance of CQL is lower than all the other HiGoC variants (above $90.0$). It is because the hierarchical planner is less greedy than the end-to-end policy. In Fig.~\ref{fig:ra-vis}, we show a sampled trajectory of the trained agent passing through a roundabout in the local map with the variant ``$7$-step HiGoC-$0.4$s''. When the ego car in red is entering the roundabout, it first yields to the two green obstacle vehicles. Afterwards, it enters the roundabout and navigates through it successfully. When the ego vehicle is equipped with a non-hierarchical offline RL policy, it is very difficult for the ego car to slow down and yield to obstacles at the entrance of the roundabout. In contrast, even though the goals sampled by the CVAE are vague and twisted, they are still informative enough to guide the planner to be less aggressive and greedy. 

\subsubsection{Different Dataset Quality}

We now compare the performance of HiGoC when trained on datasets with different levels of quality. When trained on the ``Medium'' dataset, HiGoC tends to achieve score that is closer to the corresponding expert. Since the data-collecting agent for the ``Medium'' dataset is controlled by an early-stopped policy, its policy is suboptimal with larger variance. Hence, the ``Medium'' dataset tends to cover larger state and action space. In contrast, the state and action distribution in the ``Expert'' dataset is much concentrated. The offline reinforcement learning agent is more suited in the settings where we have larger data coverage~\cite{bridging2021}. Therefore, it is reasonable that the agent is able to reach closer performance to the ``Medium'' level demonstration than the ``Expert'' level one.

% Also, the learning curves of all the variants trained on the ``Expert'' dataset converge faster than those on the ``Medium'' dataset (Fig.~\ref{fig:lc}). We can again attribute the faster convergence in the ``Expert'' dataset to the concentrated data distribution. Intuitively, the agent can quickly recognize the patterns within these similar samples from the ``Expert'' dataset, and hence converges to the highest possible score fast. The agent finds the direction to optimize its policy and value function at a slower rate on the ``Medium'' dataset. This is especially the case for CQL which has no hierarchical structure as shown in Fig.~\ref{fig:lc}(a). In contrast, the HiGoC variants improve much faster at early stages of training.

\subsubsection{Robust to Distributional Shift}
In Table.~\ref{tab:results}, the normalized scores of ``HiGoC-no noise'' are consistently lower than ``HiGoC-$0.4$s'' in almost all the settings. It indicates that our method to increase the robustness against distributional shift is effective. It is worth noting that ``HiGoC-no noise'' still outperforms CQL on both datasets, demonstrating that the hierarchical structure is beneficial to solve the overall task. 

\subsubsection{Different Goal Planning Look-ahead}

The overall planner is essentially a receding horizon controller, i.e., model predictive control (MPC).
The optimization in Eqn.~\ref{eq:uncon-opt} is solved by CEM online during the test time, and the computing time is within $100$ms which is sufficient for the framework to plan at a frequency of $10$Hz.
From control theory, if we have a precise dynamic model that can perfectly predict the behavior of the environment, the performance of the controller improves as the look-ahead horizon $H$ increases~\cite{mpc}.
In our high-level goal planner, the value function is equivalent to a prediction function estimating future cumulative rewards from the present state. Ideally, the trained agent should have better performance with longer $H$. 
However, we observed that the normalized score peaks at $7$-step look-ahead with $0.4$s goal-sampling period. The normalized score began to drop when further increasing $H$. If a goal-sampling period of $2.0$s is selected, the highest normalized score is achieved with $H=1$. It is because the learned value function becomes less accurate when predicting the far-away future. There is a trade-off between prediction accuracy and long-term planning capability. 

\subsubsection{Comparison with IRIS}
We compare HiGoC against IRIS with different look-ahead horizons. Similar to ours, their high-level planner also selects a reference goal for the low-level goal-conditioned policy. The main difference is that IRIS chooses the optimal goal as the one with the highest expected future return in the original task MDP. In contrast, HiGoC chooses the sub-goal by finding the optimal sub-goal sequence over the look-ahead horizon, where the objective function is defined based on the value functions of the short-term goal-reaching MDPs between consecutive look-ahead time steps. Also, the low-level goal reaching policy of IRIS is learned with imitation learning, whereas ours is trained with offline RL. We ran IRIS with different look-ahead horizons, i.e., the number of timesteps between the current state and the goal point. As shown in Tab.~\ref{tab:results}, HiGoC consistently outperforms IRIS under different look-ahead horizons. Two factors contribute to this performance gain. Firstly, IRIS requires accurate estimation of the value function for the original task MDP, which is difficult for temporally extended tasks, especially under the offline setting. In contrast, HiGoC composes long-term behavior online based on the value functions of the short-term goal-reaching MDPs, which are easier to estimate via offline learning. Secondly, the low-level goal-reaching policy of IRIS is trained by imitation learning instead of offline RL. The offline RL agent is able to compose behavior that is better than the demonstration from the offline dataset.

\subsection{A More Realistic Driving Scene} \label{sec:big-map}

\begin{table}[b]
    \centering
  \begin{adjustbox}{max width=\textwidth}
    \begin{tabular}{l|c|c|c}
        \toprule
        \textbf{Dataset} & \textbf{Variants} & \textbf{Normalized Score} & \textbf{Collision Rate} \\
         \midrule
         \multirow{4}{4em}{\textbf{Medium}}
         & HiGoC-$7$step & $54.2\pm9.7$ & $0.32$ \\
         & HiGoC-$1$step & $45.5\pm8.5$ & $0.38$ \\
         & IRIS-$7$step & $43.3\pm7.1$ & $0.40$ \\
         & CQL     & $31.1\pm8.3$  & $0.45$ \\
         \midrule
         \midrule
         \multirow{4}{4em}{\textbf{Expert}} 
         & HiGoC-$7$step & $61.6\pm8.4$ & $0.23$ \\
         & HiGoC-$1$step & $47.5\pm7.9$ & $0.32$ \\ 
         & IRIS-$7$step & $48.2\pm8.7$ & $0.33$ \\
         & CQL     & $37.4\pm6.2$  & $0.36$ \\
         \bottomrule
    \end{tabular}
    \end{adjustbox}
    \caption{The performance in the whole town map.}
    \label{tab:town-results}
    % \vspace{-0.3cm}
\end{table}

To further evaluate the ability of HiGoC in learning driving policies for more general and realistic scenes, we experiment HiGoC with a much larger driving scene in the CARLA simulator, the whole ``Town03'' map. We compare two settings with different look-ahead horizons, i.e., ``7-step HiGoC-$0.4$s'' and ``1 step HiGoC-$0.4$s''. As shown in Tab.~\ref{tab:town-results}, both variants of HiGoC outperforms the baseline CQL, indicating the benefit of hierarchical structure in long-horizon tasks. In particular, HiGoC-$7$step has better performance. It further confirms that longer look-ahead steps can benefit the high-level strategy reasoning even in these complex tasks. Also, the performance of HiGoC is better than IRIS. It further confirms the advantage of HiGoC over IRIS in composing optimal long-term behavior. We notice that the agent reaches performance closer to the ``Expert'' than the ``Medium'' level demonstration in this complex scenario. One possible reason is that we use the same size of training dataset as the one in the local map, which is relatively small in this much larger scene. Although the ``Medium'' level dataset consists of more diverse samples, it does not have sufficient samples to cover high-reward state-action pairs in the whole town. Thus, it limits the training quality.

It is worth noting that the collision rate with our best policy is still too high for real-world application. We notice that most of the collisions are caused by the ego vehicle bumping into the rear-end of the preceding vehicle. It is because the current representation of observations only involves implicit velocity information (the historical positions of vehicles with faded color in the images)~\cite{urbandrive}, and the lack of velocity information is also magnified by the encoding process of the CVAE. Nevertheless, we are still able to reduce the collision rate with the hierarchical architecture. It indicates that the hierarchy can prevent the agent from being too greedy.

% \begin{figure}[t]
%     \centering
%     \includegraphics[width=0.3\textwidth]{figures/lc.png}
%     \caption{(a) The learning curves of all variants trained with the ``Medium'' quality dataset; (b) the learning curves of all variants trained with the ``Expert'' quality dataset. The average return converges faster in the ``Expert'' dataset where the data coverage is narrower.}
%     \label{fig:lc}
%     \vspace{-0.3cm}
% \end{figure}

Apart from the velocity information, it is also possible to further improve the performance from other aspects of representation. The current observation representation makes it necessary for the high-level planner to reason about the future road map in addition to the surrounding vehicles. In practice, it is a common practice for autonomous vehicles to have access to the map of the whole town in advance. If we leverage a representation with richer map information, it will no longer necessary for the high-level planner to forecast map change. Plus, a better representation will also enhance the capacity of the CVAE model. With a more accurate CVAE model, the sampled goal sequences are more likely to be realizable and optimal.

\begin{figure}[t]
    \centering
    \includegraphics[width=0.33\textwidth]{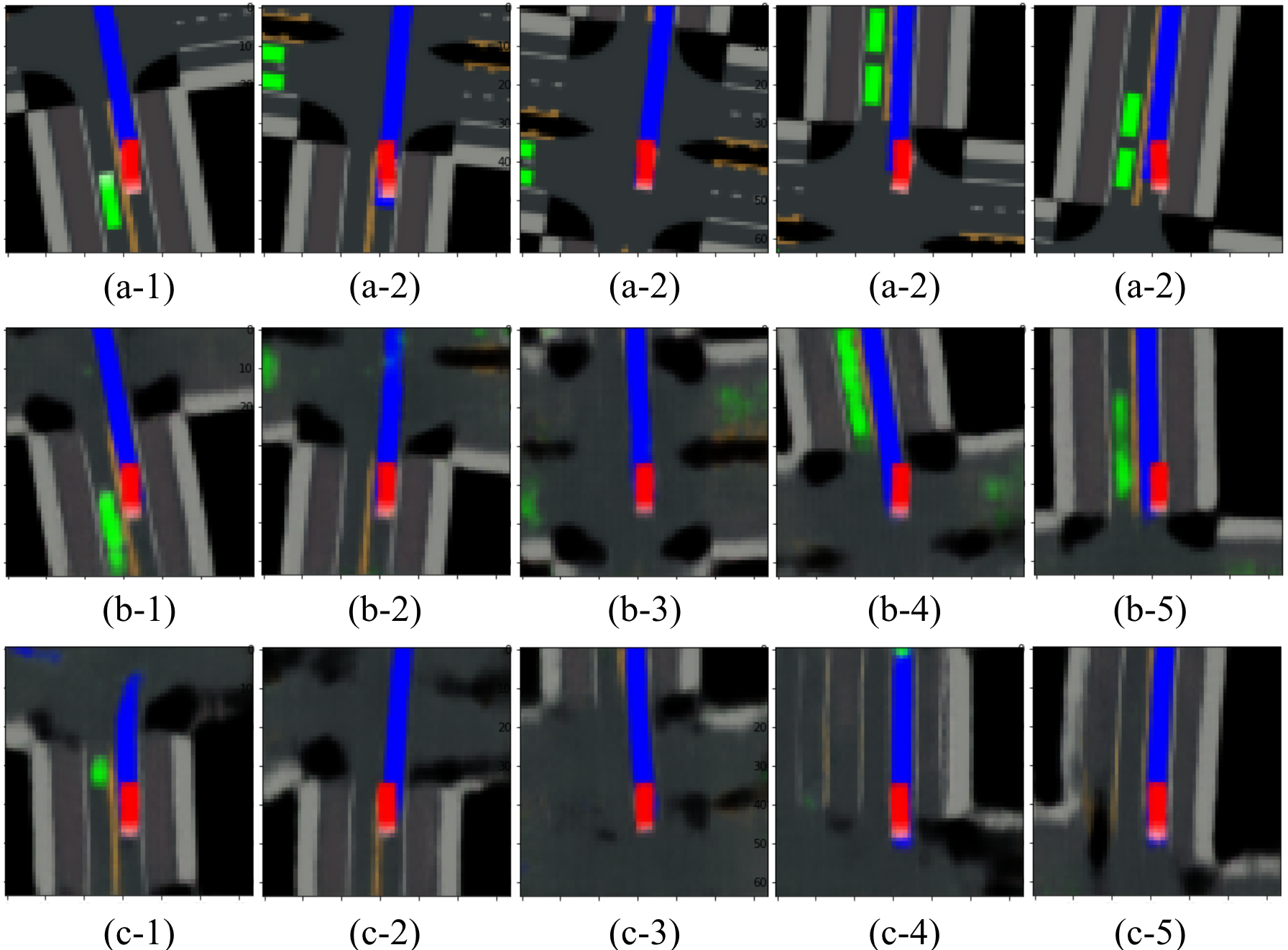}
    \caption{Visualization of a sampled trajectory. 
    (a) The actual observation images; 
    (b) The corresponding goals selected by the goal planner with CVAE; 
    (c) The corresponding goals selected by the goal planner with VAE.
    }
    \label{fig:vae-visual}
\vspace{-0.5cm}
\end{figure}

\subsection{Performance in Antmaze}

We now present the experimental results in the antmaze environments. As shown in Tab.~\ref{tab:antmaze}, HiGoC has better performance than CQL and IRIS, especially in those challenging environments (i.e., medium and large). It is because the episode length increases with the size of the maze, which makes it more critical to compose long-term optimal behavior in order to succeed in these environments. The results further verifies the advantage of HiGoC over CQL and IRIS in temporally extended tasks. In Tab. III, we also report the scores of OPAL collected from their paper~\cite{ajay2020opal}. Since OPAL is deliberately designed to learn from offline data consisting of varied and undirected multi-task behavior, they only evaluated their methods on the medium/large-diverse antmaze environments. While HiGoC is not specially designed for diverse offline data, it still achieves performance close to OPAL in those environments. 

\begin{table}[b]
\vspace{-0.3cm}
    \centering
  \begin{adjustbox}{max width=\textwidth}
    \begin{tabular}{l|c|c|c|c}
        \toprule
        Env & CQL & OPAL & IRIS & HiGoC \\
        \midrule
        umaze & $74.0$ & - & $82.6\pm4.7$ & $85.3\pm2.1$ \\
        umaze-diverse & $84.0$ & - & $89.4\pm2.4$ & $91.2\pm1.9$ \\
        medium-play & $61.2$ & - & $73.1\pm4.5$ & $81.4\pm2.4$ \\
        medium-diverse & $53.7$ & $81.1\pm3.1$ & $64.8\pm2.6$ & $79.3\pm2.5$ \\
        large-play & $15.8$ & - & $57.9\pm3.6$ & $69.1\pm2.3$ \\
        large-diverse & $14.9$ & $70.3\pm2.9$ & $43.7\pm1.3$ & $67.3\pm3.1$ \\
        \bottomrule
    \end{tabular}
    \end{adjustbox}
    \caption{Normalized scores on antmaze environments.}
    \label{tab:antmaze}
    % \vspace{-0.5cm}
\end{table}

\subsection{Discussions}
In this section, we discuss several remaining questions of interest with qualitative evaluations of the CVAE model and the learned value function. 

\subsubsection{Is CVAE better?}
Whether the CVAE model can accurately model the goal distribution is important for the performance of the high-level planner. We are then curious about if we can indeed sample better goals from the CVAE model than a VAE. In Fig.~\ref{fig:vae-visual}, we compare the goal sequences sampled from the CVAE and VAE in the same scenario. From Fig.~\ref{fig:vae-visual}(c), we can see that the goals sampled from the VAE is vaguer than those sampled from CVAE in Fig.~\ref{fig:vae-visual}(b). More importantly, the vanilla VAE model does not correctly model the surrounding vehicles in the sampled goal sequences. As shown in Fig.~\ref{fig:vae-visual}(c-4) and (c-5), the obstacle vehicles in green are almost ignored by the vanilla VAE. In contrast, the CVAE model synthesizes the surrounding agents' future locations in its sampled goals, which allows the high-level planner to gain insights about the surrounding vehicles. As a result, the planner equipped with the CVAE reaches a higher score of $61.6$ than $57.4$ of that with the VAE in the whole town map when trained using the expert demonstration. 

\subsubsection{Is the Goal-Conditioned Value Function Reliable?}
\begin{figure}[t]
    \centering
    \includegraphics[width=0.35\textwidth]{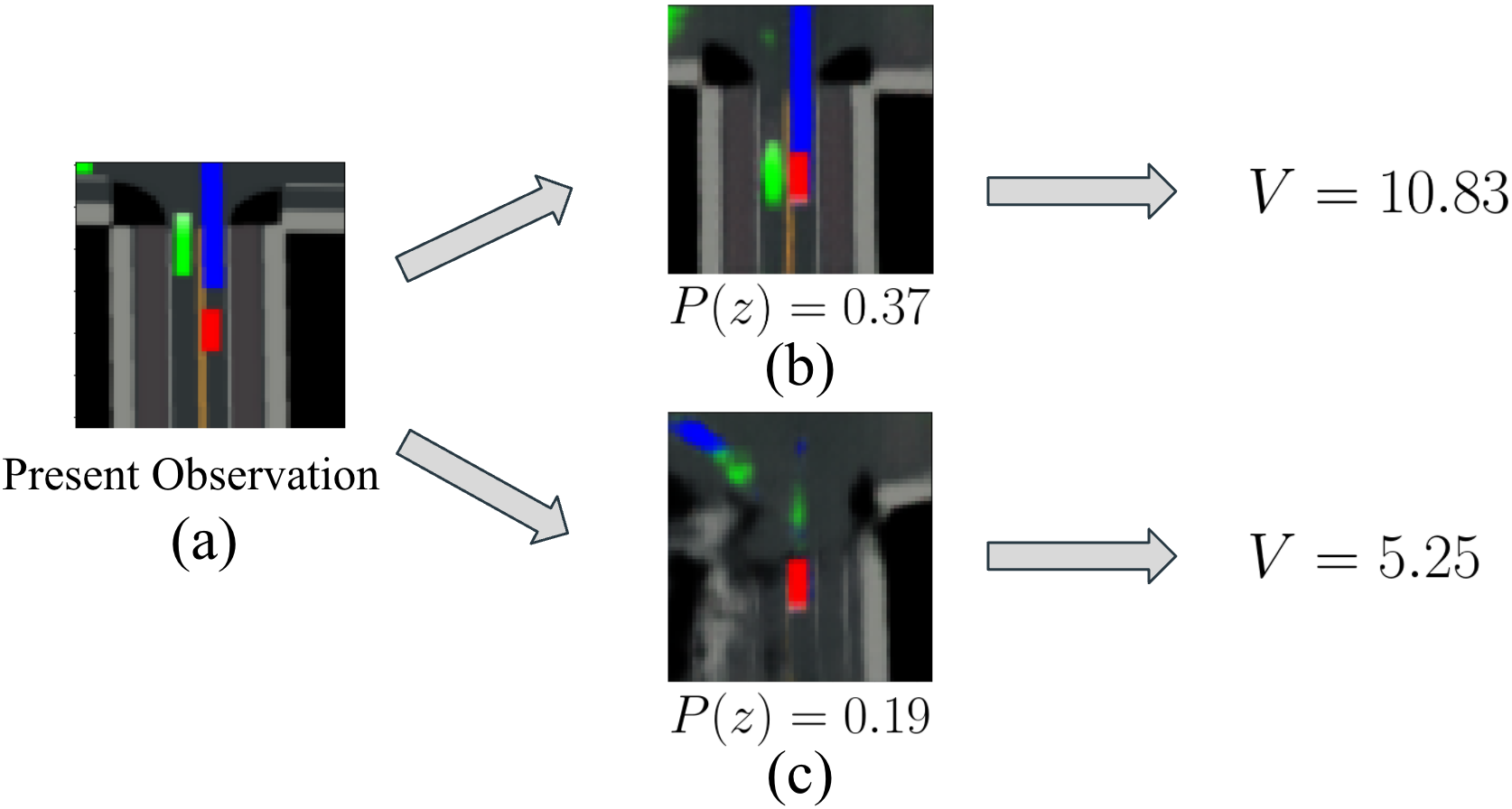}
    \caption{(a) The present observation image; (b) a high-quality goal candidate with high latent probability; (c) a low-quality goal candidate with low latent probability. 
    The value estimation is higher for the high-quality candidate.}
    \label{fig:value-func}
    \vspace{-0.3cm}
\end{figure}

In Fig.~\ref{fig:value-func}(a), we plot the present observation when the red ego car is entering an intersection. The two images in Fig.~\ref{fig:value-func}(b) and (c) are two samples of goal candidates when starting from the present observation in Fig.~\ref{fig:value-func}(a). The green obstacle car should move down, but it should not be out of sight during the goal sampling period. 
Therefore, Fig.~\ref{fig:value-func}(b) with the green obstacle in sight is more reasonable than Fig.~\ref{fig:value-func}(c).
Hence, Fig.~\ref{fig:value-func}(b) should be assigned a higher value than Fig.~\ref{fig:value-func}(c).
Also, Fig.~\ref{fig:value-func}(c) is twisted and has plenty of noise.
Compared with Fig.~\ref{fig:value-func}(b), it has lower probability in the latent space when it is encoded by the CVAE. 
A low probability means that the image is less likely to be a valid goal in the original image manifold.
As shown in Fig.~\ref{fig:value-func}, the estimated value of Fig.~\ref{fig:value-func}(b) is $10.83$ whereas $5.25$ for Fig.~\ref{fig:value-func}(c).
Hence, the learned value function can make reliable estimations on the quality of goal candidates. 

\section{Conclusion}

We propose a hierarchical planning framework through goal-conditioned offline reinforcement learning for tasks with extended temporal duration. The low-level policy is trained by offline RL in a goal-conditioned setting, which control the agent to achieve short-term sub-goals.
The offline training is improved by a perturbed goal sampling process to deal with distributional shift. The high-level goal planner takes advantage of model-based methods by composing behavior, and solves an optimization problem based on the low-level value function for long-term strategy. The proposed framework is empirically proved to be more suitable for temporally extended tasks than regular offline RL without hierarchy. 
The offline training strategy improves the robustness to distributional shift.


%%%%%%%%%%%%%%%%%%%%%%%%%%%%%%%%%%%%%%%%%%%%%%%%%%%%%%%%%%%%%%%%%%%%%%%%%%%%%%%%

%%%%%%%%%%%%%%%%%%%%%%%%%%%%%%%%%%%%%%%%%%%%%%%%%%%%%%%%%%%%%%%%%%%%%%%%%%%%%%%%

%%%%%%%%%%%%%%%%%%%%%%%%%%%%%%%%%%%%%%%%%%%%%%%%%%%%%%%%%%%%%%%%%%%%%%%%%%%%%%%%



% \section*{ACKNOWLEDGMENT}

% The preferred spelling of the word ÒacknowledgmentÓ in America is without an ÒeÓ after the ÒgÓ.

%%%%%%%%%%%%%%%%%%%%%%%%%%%%%%%%%%%%%%%%%%%%%%%%%%%%%%%%%%%%%%%%%%%%%%%%%%%%%%%%

\bibliographystyle{ieeetr}
\bibliography{references.bib}

\end{document}